\documentclass[letterpaper, 10 pt, conference]{ieeeconf}

\IEEEoverridecommandlockouts                              % This command is only needed if 
\usepackage{cite}
\usepackage{listings}
\usepackage{float}
\usepackage[dvipdfmx]{graphicx}
\usepackage{amssymb}
\usepackage{latexsym}
\usepackage{amsfonts}
\usepackage{url}
\usepackage{algorithm}
\usepackage{algpseudocode}
\usepackage{amsmath}

\onecolumn

\begin{document}

\newcommand{\ddslam}{{D{\&}D-SLAM}}

\newcommand{\tabC}{
\begin{table*}[t]
  \centering
  \caption{Relationship to Prior Work.}\label{tab:C}
  \begin{tabular}{|l|c|c|c|c|}
    \hline
\multirow{2}{*}{Approaches} & \multicolumn{2}{c|}{Static Scenes} & \multicolumn{2}{c|}{Dynamic A14-15} \\
\cline{2-5}
&  A2 & A3 & A4 & A5 \\
\hline
\hline
conventional SLAM & A22 & A23 & A24 & A25 \\
    \hline
Dynamic SLAM (e.g., \cite{DynamicSLAM}) & \cm & \cm &  & \\
    \hline
Thermal SLAM (e.g., \cite{9804793})& \cm &  & \cm & \\
    \hline
\ddslam ~(Ours.) & A52 & A53 & A54 & A55 \\
    \hline
  \end{tabular}
\end{table*}
}

\newcommand{\cm}{\checkmark}

\renewcommand{\tabC}{
\begin{table}[t]
  \centering
  \caption{Relationship to Prior Work.}\label{tab:C}
  \begin{tabular}{|l|c|c|c|c|}
    \hline
Scene Type & \multicolumn{2}{c|}{Well-Lit} & \multicolumn{2}{c|}{Low-Light} \\
\hline
Landmark Type & S & D & S & D \\
\hline
\hline
conventional SLAM & \cm & & & \\
    \hline
Dynamic SLAM (e.g., \cite{DynamicSLAM}) & \cm & \cm &  & \\
    \hline
Thermal SLAM (e.g., \cite{9804793})& \cm &  & \cm & \\
    \hline
\ddslam (Ours.) & \cm & \cm & \cm & \cm \\
    \hline
  \end{tabular}\\
S: Static Landmark. D: Dynamic Landmark.
\end{table}
}

\newcommand{\FIG}[3]{
\begin{minipage}[b]{#1cm}
\begin{center}
\includegraphics[width=#1cm]{#2}\\
{\scriptsize #3}
\end{center}
\end{minipage}
}

\newcommand{\FIGU}[3]{
\begin{minipage}[b]{#1cm}
\begin{center}
\includegraphics[width=#1cm,angle=180]{#2}\\
{\scriptsize #3}
\end{center}
\end{minipage}
}

\newcommand{\FIGm}[3]{
\begin{minipage}[b]{#1cm}
\begin{center}
\includegraphics[width=#1cm]{#2}\\
{\scriptsize #3}
\end{center}
\end{minipage}
}

\newcommand{\FIGR}[3]{
\begin{minipage}[b]{#1cm}
\begin{center}
\includegraphics[angle=-90,width=#1cm]{#2}
\\
{\scriptsize #3}
\vspace*{1mm}
\end{center}
\end{minipage}
}

\newcommand{\FIGRpng}[5]{
\begin{minipage}[b]{#1cm}
\begin{center}
\includegraphics[bb=0 0 #4 #5, angle=-90,clip,width=#1cm]{#2}\vspace*{1mm}
\\
{\scriptsize #3}
\vspace*{1mm}
\end{center}
\end{minipage}
}

\newcommand{\FIGCpng}[5]{
\begin{minipage}[b]{#1cm}
\begin{center}
\includegraphics[bb=0 0 #4 #5, angle=90,clip,width=#1cm]{#2}\vspace*{1mm}
\\
{\scriptsize #3}
\vspace*{1mm}
\end{center}
\end{minipage}
}

\newcommand{\FIGpng}[5]{
\begin{minipage}[b]{#1cm}
\begin{center}
\includegraphics[bb=0 0 #4 #5, clip, width=#1cm]{#2}\vspace*{-1mm}\\
{\scriptsize #3}
\vspace*{1mm}
\end{center}
\end{minipage}
}

\newcommand{\FIGtpng}[5]{
\begin{minipage}[t]{#1cm}
\begin{center}
\includegraphics[bb=0 0 #4 #5, clip,width=#1cm]{#2}\vspace*{1mm}
\\
{\scriptsize #3}
\vspace*{1mm}
\end{center}
\end{minipage}
}

\newcommand{\FIGRt}[3]{
\begin{minipage}[t]{#1cm}
\begin{center}
\includegraphics[angle=-90,clip,width=#1cm]{#2}\vspace*{1mm}
\\
{\scriptsize #3}
\vspace*{1mm}
\end{center}
\end{minipage}
}

\newcommand{\FIGRm}[3]{
\begin{minipage}[b]{#1cm}
\begin{center}
\includegraphics[angle=-90,clip,width=#1cm]{#2}\vspace*{0mm}
\\
{\scriptsize #3}
\vspace*{1mm}
\end{center}
\end{minipage}
}

\newcommand{\FIGC}[5]{
\begin{minipage}[b]{#1cm}
\begin{center}
\includegraphics[width=#2cm,height=#3cm]{#4}~$\Longrightarrow$\vspace*{0mm}
\\
{\scriptsize #5}
\vspace*{8mm}
\end{center}
\end{minipage}
}

\newcommand{\FIGf}[3]{
\begin{minipage}[b]{#1cm}
\begin{center}
\fbox{\includegraphics[width=#1cm]{#2}}\vspace*{0.5mm}\\
{\scriptsize #3}
\end{center}
\end{minipage}
}

% \bfseries
% \sffamily

\newcommand{\RGB}[2]{$f_{\theta^{\text{RGB}}}(x_\text{#2}^\text{#1})$}
\newcommand{\T}[2]{$f_{\theta^{\text{T}}}(x_\text{#2}^\text{#1})$}
\newcommand{\Fusion}{$f_{\theta^{\text{RGB}}}(x_\text{Bright}^\text{RGB})$+$f_{\theta^{\text{RGB}}\theta^{\text{T}}}(x_\text{Dark}^\text{T})$}

\newcommand{\tabA}{
\begin{table}[t]
    \centering
    \caption{Tracking Accuracy}
    \label{tab:A}
    \begin{tabular}{lrrrrrr}
        \hline
        Tracker & MOTA & IDF1 & FP & TP & FN \\
        \hline \hline
BRIGHT & 99.8 & 100.0 & 5 & 8273 & 3 \\
        \hline
        RGB      & 26.3  & 42.6  & 33  & 2007 & 5371 \\
        T    & 89.4  & 95.0  & 631 & 8053 & 225   \\
        FUSION   & 87.9  & 94.8  & 661 & 8058 & 220   \\
        \hline
    \end{tabular}\\
{\scriptsize
BRIGHT: \RGB{T}{Bright},
RGB: \RGB{T}{Dark}, T: \T{T}{Dark}, FUSION: 
$f_{\theta^{\text{RGB}}\theta^{\text{T}}}(x_\text{Dark}^\text{T}, x_\text{Dark}^\text{T})$.
}
\end{table}
}

\newcommand{\tabB}{
\begin{table}[t]
    \centering
    \caption{Trajectory Prediction Performance}
    \label{tab:B}
    \begin{tabular}{lrr}
        Tracker & HOTA & IDSW \\
        \hline \hline
BRIGHT & 99.9 & 11 \\
\hline
        RGB      & 27.1  & 34  \\
        T    & 90.4  & 19  \\
        FUSION   & 90.1  & 119  \\
        \hline
    \end{tabular}
\end{table}
}

\newcommand{\figD}{
\begin{figure}[t]
\centering
\hspace*{5mm}
\FIG{6}{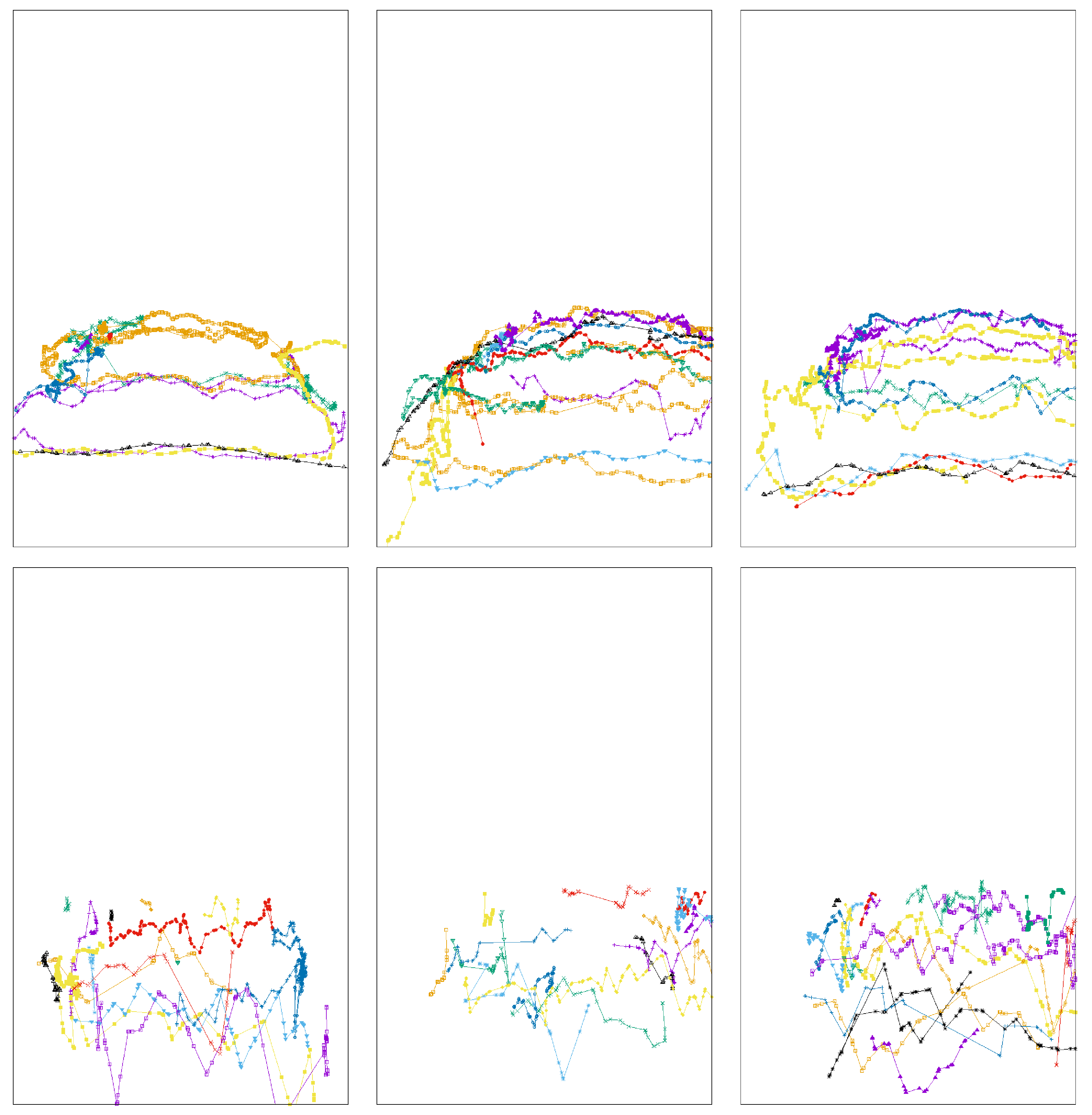}{}%
\caption{
Predicted human trajectories by ByteTrack. The top row shows $f_{\theta^\text{T}}$ trained on T images, while the bottom row shows $f_{\theta^\text{RGB}}$ trained on RGB images.
The three different columns indicate different input image sequences, and different colors represent different person IDs.
}
\label{fig:D}
\end{figure}
}

\newcommand{\figE}{
\begin{figure}[t]
\centering
\hspace*{5mm}
\FIG{6}{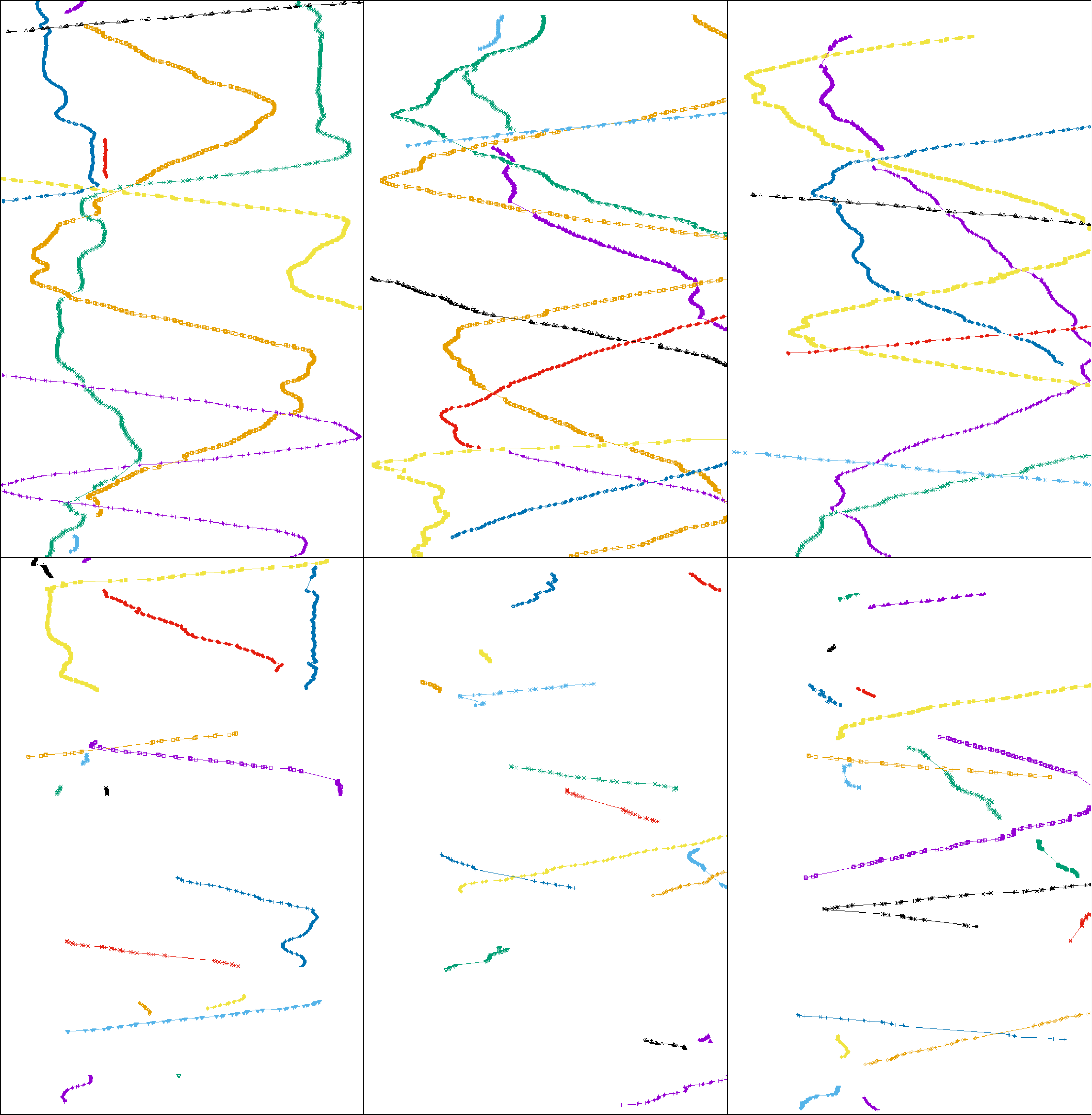}{}%
\caption{
Human movement trajectories x-t predicted by ByteTrack. The top row shows $f_{\theta^\text{T}}$ trained on T images, while the bottom row shows $f_{\theta^\text{RGB}}$ trained on RGB images.
}
\label{fig:E}
\end{figure}
}

\newcommand{\figDE}{
\begin{figure}[t]
\centering
\FIG{4}{m.eps}{(a)}\hspace*{-2mm}%
\FIG{4}{xt.eps}{(b)}%
\caption{
Visualization of $f_{\theta^\text{T}}(x^\text{T}_\text{Bright})$ (top) and $f_{\theta^\text{RGB}}(x^\text{T}_\text{Bright})$ (bottom).
The three-dimensional $x$-$y$-$t$ trajectories are projected onto the xy plane and xt plane, shown in panels (a) and (b), respectively.
In each panel, the three different columns represent different input image sequences, and different colors indicate different person IDs.
}
\vspace*{-5mm}
\label{fig:DE}
\end{figure}
}

\newcommand{\figC}{
\begin{figure}[t]
\centering
\hspace*{5mm}
\FIG{8}{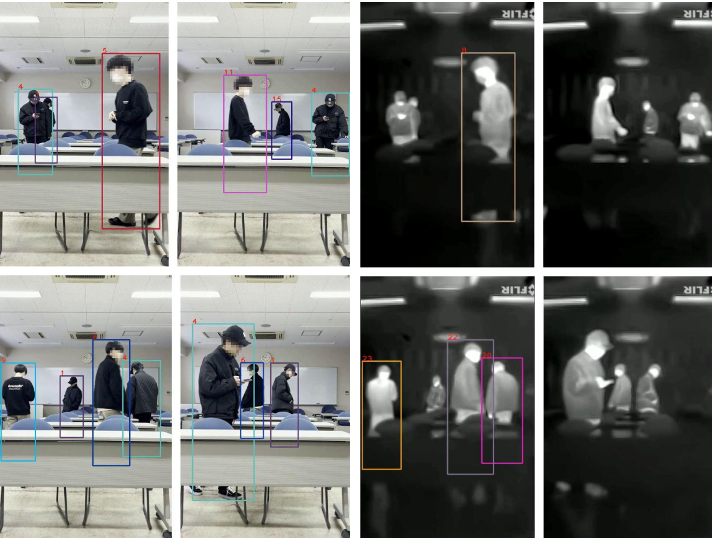}{}
\caption{
Output of the RGB tracker for the RGB image (left) and the T image (right).
}
\label{fig:C}
\end{figure}
}

\newcommand{\figI}{
\begin{figure}[t]
\centering
\hspace*{5mm}\FIG{8}{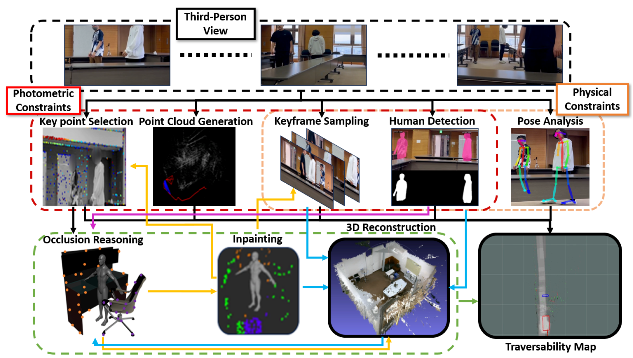}{}~\vspace*{0mm}\\
\caption{
Block diagram of the HO3-SLAM framework (adapted from \cite{DBLP:conf/iros/Liang024}). 
% All modules are interconnected via ROS (Robot Operating System). The PHYS module comprises DSO (Direct Sparse Odometry) and is responsible for generating point clouds utilized by other modules. Within the PHOT module, a Human-Object Occlusion Ordering Algorithm is employed to extract occlusion ordering information, which is then combined with point cloud coordinates derived from Detectron2 human masks. Additionally, the Walk2Map++ module utilizes human pose estimation to predict human distance from the camera and estimate traversable regions. These traversability maps are visualized using the rviz visualizer. In the traversability map image, the red box represent the estimated human location, hence the traversable region. The grey path indicates the traversable region, which has been walked by the human. The grey path is inprinted by the red boxes. 
}
\label{fig:I}
\end{figure}
}

\newcommand{\figJ}{
\begin{figure}[t]
\centering
\hspace*{5mm}\FIG{8}{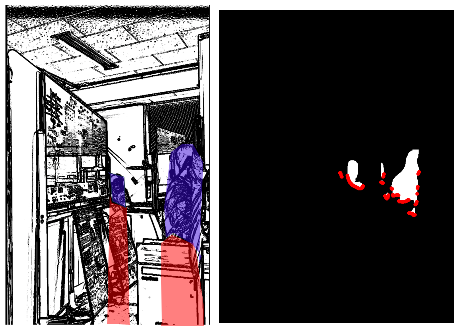}{}~\vspace*{0mm}\\
\caption{
Human-Object Interaction (HOI) Features.
Left: A rough sketch of the view, showing the human region (blue) and occluder objects (red).
Right: The detected human region (white) and HOI feature points (red) in the thermal image.
Although the occluder objects are barely visible in the thermal image, their characteristics can still be inferred from the arrangement of the boundary features.
}
\label{fig:J}
\end{figure}
}

\newcommand{\figA}{
\begin{figure}[t]
\centering
\hspace*{5mm}
\FIG{8}{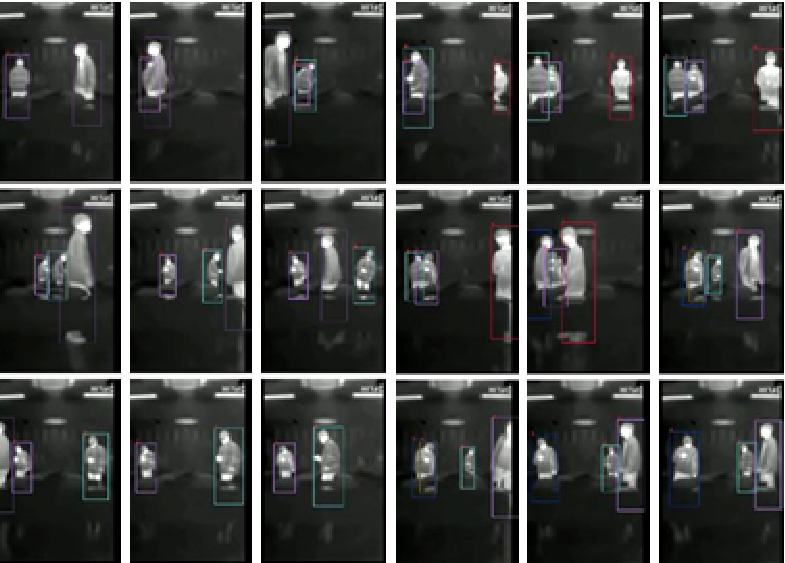}{}\vspace*{0mm}\\
\caption{
Improved tracking results from the proposed method.
}
\label{fig:A}
\end{figure}
}

\newcommand{\figB}{
\begin{figure}[t]
  \centering
\hspace*{5mm}%
\FIG{2.3}{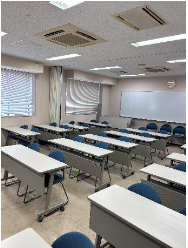}{}%
\hspace*{-1mm}%
\FIG{5.3}{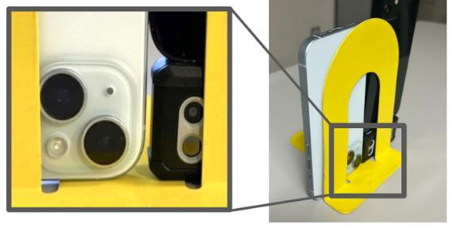}{}~\vspace*{0mm}\\
\caption{
Crowded scenes and RGB-T cooperative vision.
}
\label{fig:B}
\end{figure}
}

\newcommand{\figG}{
\begin{figure}[t]
  \centering
\FIG{8}{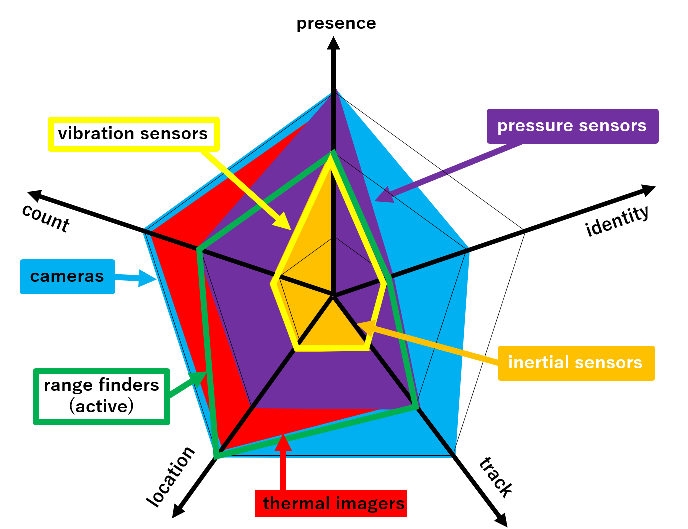}{}~\\
\caption{Human sensing capabilities of major passive sensors (and active sonar \cite{DBLP:conf/iros/ZhaTH97}), following the human sensing surveys in \cite{HumanSensing2010} and \cite{HumanSensing2023}.}
\label{fig:G}
\end{figure}
}

\newcommand{\figH}{
\begin{figure}[t]
  \centering
\FIG{8}{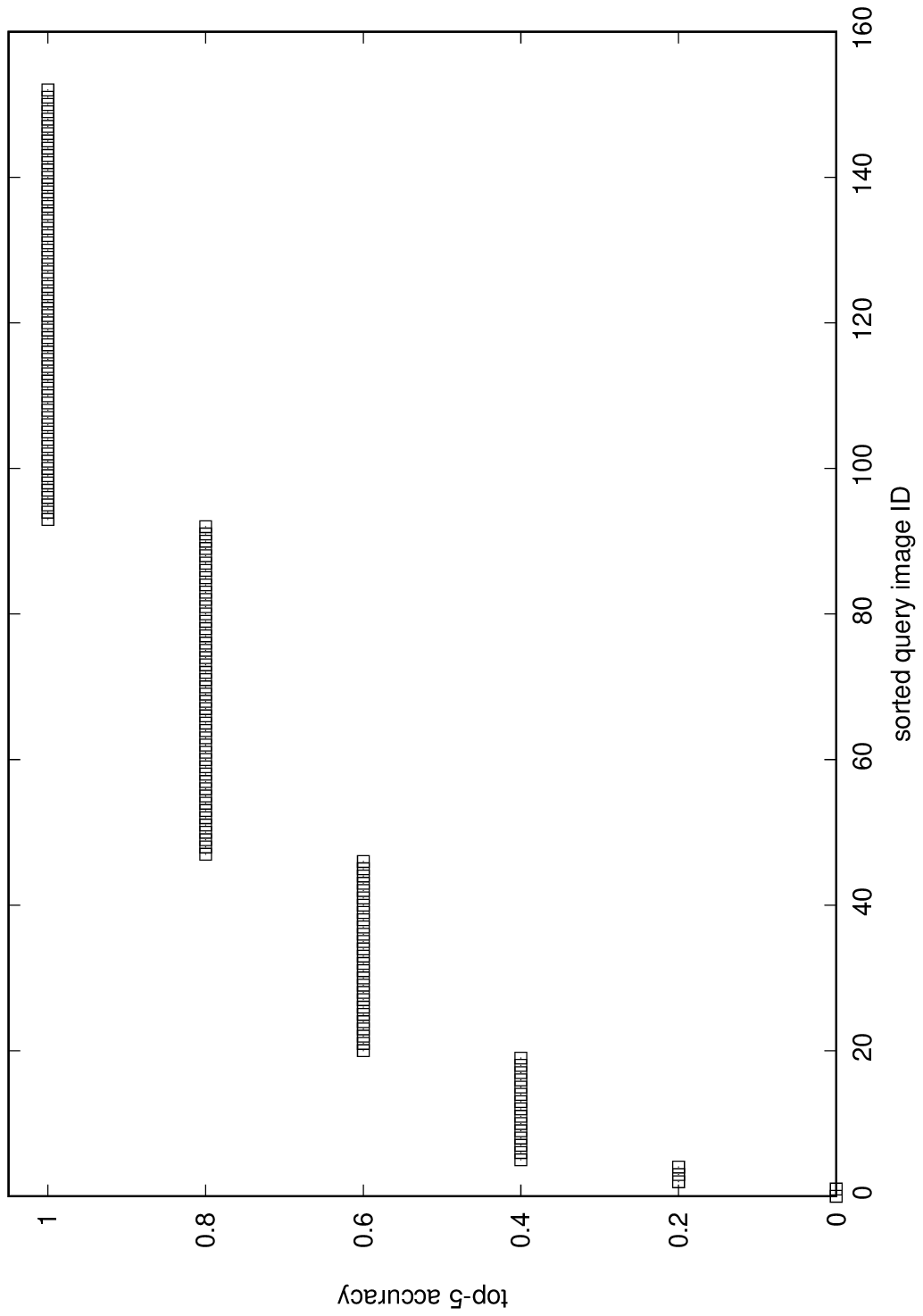}{}
\caption{Loop closure detection performance. 
Vertical axis: Top-5 accuracy $(X/5, X\in\{0, 1, 2, 3, 4, 5\})$.
Horizontal axis: Sorted query image ID.
}\label{fig:H}
\end{figure}
}

\newcommand{\figF}{
\begin{figure}[t]
  \centering
\hspace*{5mm}%
\FIG{2.5}{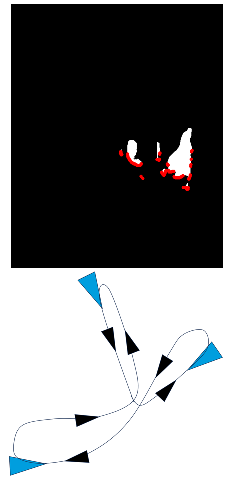}{~\\(a)}%
\FIG{5.5}{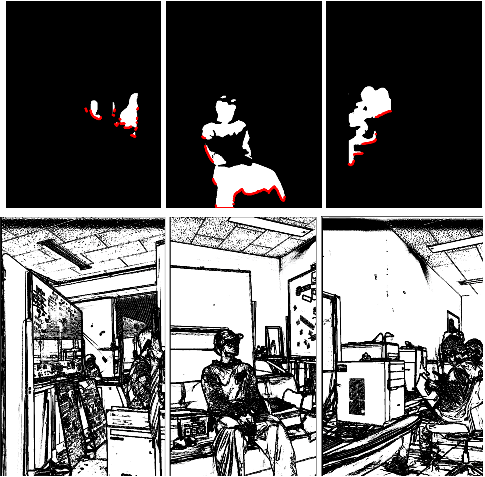}{~\\(b)}
\caption{
Loop closing experiments.
(a) A query image (top), and a rough sketch of a looped trajectory with three viewpoints along the trajectory (bottom).
(b) Thermal images (top) and rough scene sketches (bottom) acquired at those three viewpoints.
}
\label{fig:F}
\end{figure}
}

\title{
Dynamic-Dark SLAM:\\
RGB-Thermal Cooperative Robot Vision Strategy for Multi-Person Tracking in Both Well-Lit and Low-Light Scenes
}

\author{
Tatsuro Sakai, Kanji Tanaka, Yuki Minase, Jonathan Tay Yu Liang, Muhammad Adil Luqman, Daiki Iwata\\
University of Fukui\\
3-9-1, bunkyo, fukui, fukui, Japan\\
  {\tt tnkknj@u-fukui.ac.jp}\\
}

\maketitle

\section*{\centering Abstract}
\textit{
In robot vision, thermal cameras hold great potential for recognizing humans even in complete darkness. However, their application to multi-person tracking (MPT) has been limited due to data scarcity and the inherent difficulty of distinguishing individuals. In this study, we propose a cooperative MPT system that utilizes co-located RGB and thermal cameras, where pseudo-annotations (bounding boxes and person IDs) are used to train both RGB and thermal trackers. Evaluation experiments demonstrate that the thermal tracker performs robustly in both bright and dark environments. Moreover, the results suggest that a tracker-switching strategy --- guided by a binary brightness classifier --- is more effective for information integration than a tracker-fusion approach. 
As an application example, we present an image change pattern recognition (ICPR) method, the ``human-as-landmark,'' which combines two key properties: the thermal recognizability of humans in dark environments and the rich landmark characteristics---appearance, geometry, and semantics---of static objects (occluders). Whereas conventional SLAM focuses on mapping static landmarks in well-lit environments, the present study takes a first step toward a new Human-Only SLAM paradigm, ``\ddslam,'' which aims to map even dynamic landmarks in complete darkness.
Additionally, this study demonstrates that knowledge transfer between thermal and depth modalities enables reliable person tracking using low-resolution 3D LiDAR data without RGB input, contributing an important advance toward cross-robot SLAM systems.
}

\section{Introduction}

Thermal cameras (T-cameras) hold great potential for person detection and tracking in dark or low-light scenes, representing a promising new application in robot vision. Despite the increasing demand for advanced vision systems that consider privacy, such as those used in nighttime festivals and security patrols, T-vision remains underutilized, particularly in its application to multi-person tracking (MPT). MPT is a crucial technology in robot vision and is employed in various applications, including user tracking~\cite{Escort}, patrolling~\cite{Patrol}, dynamic SLAM~\cite{DynamicSLAM}, structure-from-occlusion~\cite{Structure-From-Occlusion}, and traversability analysis~\cite{TraversabilityAnalysis}. Research on MPT has primarily been based on RGB vision~\cite{RGBMPT}, while challenges specific to T-vision remain largely unaddressed.

One of the major obstacles in applying T-vision to MPT is the lack of annotated datasets (bounding boxes + person IDs), which makes supervised training of a T-tracker infeasible. These datasets are typically represented as time-series pairs of bounding boxes and person IDs. However, most existing datasets focus on RGB vision, and thermal image datasets are not yet well established. Furthermore, person ID identification in thermal images is difficult and requires high-cost annotation efforts to achieve accuracy. 
As a result, applying tracking techniques designed for RGB vision to T-vision leads to a decrease in performance (Fig. \ref{fig:C}).
Notably, the patterns in thermal images heavily depend on factors such as clothing, making them more complex than traditional RGB image patterns. Due to this dependency, generalizing thermal image datasets to arbitrary environments is difficult, necessitating the development of T-trackers that do not rely on annotated datasets.

\figB

In this study, we propose a novel cooperative MPT system that utilizes an RGB camera and a thermal (T) camera placed in the same pose, enabling effective tracking in both bright and dark conditions. The key idea is to train a T-tracker using knowledge transfer from an RGB-tracker in bright conditions. 
Among all possible combinations of
(1) two scene types (Bright/Dark),
(2) two modality types (RGB/T),
(3) three tracker types (RGB/T/RGBT tracker),
we exclude those that are clearly ineffective or redundant, leaving four viable combinations, as discussed in Section \ref{sec:exp}, which we then evaluate experimentally.
The results demonstrate that the T-tracker achieves remarkable performance in both bright and dark scenes. Furthermore, our findings suggest that a tracker-switching approach using a binary brightness classifier is more suitable than a tracker-fusion approach for integrating the two trackers.

\tabC

As an application example, we propose \textit{\ddslam}, a novel SLAM framework that leverages dynamic landmarks in complete darkness, to recognize, understand, and reconstruct people, occluder objects, and traversable areas.
This framework is motivated by recent advances in RGB and thermal-based human activity recognition \cite{10802830}---particularly in identifying human-object interactions, such as sitting on a chair or leaning on a desk.
Building on this insight, we introduce ``human-as-landmark,'' an image change pattern recognition (ICPR) method that exploits two key properties: 
(1) the thermal visibility of humans in the dark, and 
(2) the rich landmark features---appearance, geometry, and semantics---of static objects involved in human interaction.
We experimentally demonstrate that even a simple implementation of human-as-landmark improves loop closure detection, a task where conventional methods frequently fail due to false negatives.

In addition to the RGB-T cooperative vision, this study also explores person tracking using Thermal (TIR) and Depth images without RGB input.  
By leveraging knowledge transfer across these different sensor modalities, the approach achieves reliable multi-person tracking even with low-resolution depth images derived from 3D LiDAR sensors.  
This extension broadens the applicability of cooperative vision systems and represents an important step toward cross-robot SLAM.  
Details and experimental results of this Thermal-Depth cooperative tracking are presented in Section~\ref{sec:td_tracking}.

\figC

\section{Related Work}

This study is related to several areas of research in robotic vision.
In the field of robotics, research on multi-person tracking (MPT) has progressed using laser-based \cite{RoboticsLaserTracking} and vision-based \cite{RoboticsVisionTracking} methods, with the study of SLAM considering interactions between dynamic and static objects \cite{DynamicSLAM} being the most relevant to our work.
However, most existing vision-based methods focus on RGB vision, leaving challenges specific to thermal (T) vision largely unresolved.

In the field of computer vision, studies on multi-sensor information fusion \cite{MultisensorFusion}, 
such as RGB-LiDAR fusion, 
as well as research on teacher-student knowledge transfer \cite{KnowledgeTransfer}, have been conducted.
However, many of these studies rely on predefined cooperative schemes \cite{PredefinedCooperation} where the teacher and student sensors are used complementarily at the same time, with little exploration of diverse cooperative strategies.

RGB-T fusion tracking has been studied in both non-deep approaches  
\cite{DBLP:journals/mva/ConaireOS08, DBLP:conf/cvpr/BirchfieldR05, DBLP:conf/fusion/WuBCBL11, DBLP:journals/chinaf/LiuS12}  
and deep approaches, including adaptive fusion \cite{DBLP:journals/tip/LiCHLTL16},  
sparse fusion representation \cite{DBLP:journals/tsmc/LiSWZT17},  
cross-modal ranking \cite{DBLP:conf/eccv/LiZHTW18},  
and graph models \cite{DBLP:conf/mm/LiZLZT17}, as well as benchmark studies \cite{DBLP:journals/pr/LiLLZT19}.  
The approach of applying an RGB tracker \cite{9010649} to the RGB-T tracking problem was explored in \cite{DBLP:journals/corr/abs-1908-11714}.  
Our approach is based on the state-of-the-art ByteTrack \cite{ByteTrack} tracker and differs in that it handles Dark scenes,  
where RGB trackers completely fail.

\section{Dynamic-Dark SLAM (\ddslam)}

The combination of low-light conditions and dynamic landmarks constitutes one of the most challenging scenarios for SLAM technologies.  
Most existing SLAM approaches assume the presence of some illumination (i.e., not complete darkness) and the presence of recognizable static landmarks within the scene \cite{10075065}. Accordingly, much of the literature has focused on mapping and localization based on static landmarks.

Our approach takes an opposite yet complementary perspective. We propose \textit{Dynamic-Dark-SLAM (\ddslam)}, a novel human-only SLAM framework that diverges from conventional bearing-range \cite{ramezani2020online}, range-only \cite{song2019uwb} or bearing-only \cite{bj2017redesign} SLAM paradigms by using dynamic objects—specifically, humans—as landmarks in completely dark environments. Since RGB cameras are ineffective in total darkness, our method instead leverages thermal (T) cameras.

\subsection{Challenges}

This paradigm shift introduces several unique challenges:

Human sensing suffers from high noise levels, large intra-class variation and non-rigid appearance (e.g., clothing, posture, accessories), and inter-person similarity, making re-identification difficult.

Landmark dynamics are significant because human landmarks frequently appear, disappear, and move. This movement compromises the 100\% precision \cite{zhang2023binary} required for loop closure, risking catastrophic mapping errors. It is noteworthy that the HO3 landmarks are intentionally transient—they may appear, disappear, or move—and are used primarily for loop closure rather than for persistent long-term map storage.

Sensor limitations include difficulty in recognizing appearance, geometry, and semantics under thermal imaging, absence of color cues, and variability in thermal signatures due to body temperature changes.

\subsection{Related Work and Inspirations}

Our approach is inspired by two lines of prior research. First, loop-closure detection and image change detection (LCD-ICD) approaches \cite{DBLP:conf/iros/ZhaTH97} and its follow-ups \cite{DBLP:conf/iros/TakedaTN22} demonstrated that when the relative position between a landmark A and an object B remains unchanged, both can be considered static. Second, human-only SLAM works \cite{DBLP:conf/iros/Tanaka02} and \cite{DBLP:conf/iros/Liang024} treat the occlusion boundary between a human A and an occluding object B as a pseudo-static landmark feature, albeit prone to frequent false negatives.

\figI

Based on these ideas, we have developed \textit{HO3-SLAM (Human Object Occlusion Ordering SLAM)} \cite{DBLP:conf/iros/Liang024} (Fig. \ref{fig:I}), where HOI boundary points are treated as key landmark feature points for mapping and localization, with a particular focus on loop closure detection. Since this occlusion boundary represents interactions between a dynamic human and a static occluder, it combines the desirable properties of both (Fig.~\ref{fig:J}): it is stably detectable via human tracking in both RGB and T domains and is invariant in appearance, geometry, and semantics—thus serving as a reliable landmark.

\figJ

\subsection{Key Difference from HO3-SLAM}

The primary difference in \textit{\ddslam} compared to HO3-SLAM is the replacement of RGB cameras with thermal (T) cameras. This change introduces the following implications:

Human sensing with T cameras shows significant degradation in tracking and identification performance (Fig.~\ref{fig:G}) \cite{HumanSensing2023}.

Static object sensing with T cameras is challenged by noise, ambient temperature effects, and low spatial resolution, as widely noted in thermal SLAM literature \cite{9804793}.

The idea of using humans as cues for mapping, as well as the concept of mapping dynamic objects, is not entirely novel—prior work \cite{656464} and \cite{DBLP:journals/corr/abs-1301-0551} has explored these topics. However, this perspective has been scarcely explored within the context of HO3-SLAM or the broader human-as-landmark paradigm.

\subsection{Summary and Outlook}

In summary, \textit{\ddslam} is a new SLAM paradigm characterized by dark scenes and dynamic landmarks, in contrast to conventional SLAM approaches designed for well-lit and static landmarks. We emphasize the urgent need for a new class of Image Change Pattern Recognition (ICPR) techniques that can operate under Non-Instrumented, Non-Color, Non-Active(-Electromagnetic) (NICA)  
\footnote{Instrumented: using external devices such as GPS sensors, wearable sensors attached to objects or people, or multi-view sensing systems like stereo cameras. Color: refers to sensors that measure surface reflectivity in the visible or near-visible spectrum, such as RGB cameras, grayscale cameras, and multispectral sensors. Active: refers to active(-electromagnetic) sensors such as LiDAR, ToF, structured light, etc.}  
sensor modalities. The task of ICPR using NICA sensors in non-static and non-bright scenes has been explored in our previous work \cite{DBLP:conf/iros/ZhaTH97}, but the exploration of \ddslam is presented here for the first time. Thermal cameras offer a unique advantage—functioning in total darkness—but are inherently difficult to use, highlighting the “no free lunch (NFL)” principle. D\&D-SLAM demands innovation across the ICPR pipeline: image change detection, counting, tracking, localization, and (re-)identification.

\figG

\section{RGB-T Cooperative MPT}

\subsection{Formulation of MPT}

MPT is formulated as a problem where a tracker $f$ takes an image sequence $x_{1:T}$ as input and outputs a sequence of tracking results $y_{1:T}$. Here, $x_{1:T} = (x_1, \cdots, x_T)$ represents the sequence of camera images, where $x_t$ denotes the image at time $t$. The tracking result is represented as $y_t = \{y_t[i]\}_{i=1}^{I_t}$, where 
$I_t$ is the number of bounding boxes detected in the image $x_t$
and each $y_t[i]$ consists of $(y_t^\text{ID}[i], y_t^\text{BB}[i], y_t^\text{SCORE}[i])$. Specifically, $y_t^\text{ID}[i]$ is the predicted ID of the person in the $i$-th bounding box, $y_t^\text{BB}[i]$ represents the predicted location and size of the $i$-th bounding box, and $y_t^\text{SCORE}[i]$ indicates the confidence score.

Given an annotated image sequence $(x_{1:T}, y_{1:T})$, the tracker is trained as follows:
\begin{equation}
\theta = \arg\min_{\theta} \mathcal{L} ( f_{\theta}(x_{1:T}), y_{1:T} )
\end{equation}
where $\mathcal{L}$ represents the loss function that represents the error between the predicted results and the annotations.

\subsection{ByteTrack}

In this study, we propose using ByteTrack\cite{ByteTrack} as the baseline tracker, which separates detection from tracking and remains robust even under temporary full or partial occlusion. Many existing trackers discard detection results with scores below a threshold, whereas ByteTrack adopts a simple and general approach that utilizes low-score objects.
Its core module consists of an object detector that outputs bounding boxes, scores, and classes for each frame, as well as a Byte module that associates tracked object IDs with detection results. ByteTrack employs a Kalman filter to predict the position of objects in frame $t$ based on tracking information up to frame $t-1$. Then, it associates predicted bounding boxes with actual detections in frame $t$ based on Intersection over Union (IoU), matching the pairs with the highest IoU and selecting them based on a threshold.

This simple yet effective two-stage IoU matching method enables stable and fast tracking. In practice, ByteTrack demonstrated superior performance in our study. In addition to the main experiments, we conducted a comparative evaluation of three state-of-the-art trackers: OC-SORT \cite{DBLP:conf/cvpr/CaoPWKK23}, StrongSORT \cite{DBLP:journals/tmm/DuZSZSGM23}, and ByteTrack \cite{ByteTrack}. The RGB image sequence captured in a well-lit scene was used for this comparison because of the availability of high-quality ground-truth for such sequences. The results showed that ByteTrack significantly outperformed OC-SORT and slightly more stable than StrongSORT.

The baseline tracker ByteTrack was trained as follows:
\begin{equation}
\theta^\text{RGB} = \arg\min_{\theta} \mathcal{L}^\text{ByteTrack} ( f_{\theta}(x^\text{RGB}_{1:T}), y_{1:T} )
\end{equation}
where $\mathcal{L}^\text{ByteTrack}$ represents the loss function used in ByteTrack.
The training set $(x^\text{RGB}_{1:T}, y_{1:T})$
consists of five RGB image datasets: MOT17\cite{MOT17}, MOT20\cite{MOT20}, CrowdHuman\cite{CrowdHuman}, CityPersons\cite{CityPersons}, and ETHZ\cite{ETHZ}.
These datasets encompass various environments and scenarios, including urban settings, high-density crowds, occlusion scenarios, diverse weather conditions, and long-range pedestrian tracking, enhancing the model's adaptability.

\subsection{Cooperative Strategy}

An RGB camera and a thermal (T) camera were placed at the same position and synchronized. By positioning the lenses of both cameras as closely as possible, we minimized the misalignment of the fields of view, reducing coordinate errors between the RGB and thermal images. Additionally, temporal discrepancies and resolution differences were corrected after capturing the data to precisely align the coordinates between the RGB and thermal images.

The input image at time $t$ is defined as $x_t = (x_t^\text{RGB}, x_t^\text{T})$, where $x_t^\text{RGB}$ represents the RGB image, and $x_t^\text{T}$ represents the thermal image at time $t$.

The optimal coordinate transformation between the two camera coordinate systems may vary slightly depending on the environment. For instance, in a crowded environment such as that shown in Figure \ref{fig:C}, pedestrians move while avoiding obstacles, leading to motion trajectories constrained by the obstacles. Therefore, it is possible to optimize the coordinate transformation based on pedestrian movement trajectories. In this study, a small set of image pairs of the same pedestrians observed from different cameras was used as training data to adjust the coordinate transformation parameters, minimizing transformation errors. Since the experimental environment ensured that both cameras captured the same plane or had minimal parallax, we modeled the relationship between the cameras using a homography transformation.

The experiments were conducted in an indoor environment with obstacles such as desks and chairs. 
Furthermore, we evaluated the multi-person tracking performance in a scene with four pedestrians.

In the experiment, an iPhone 15 was used as the RGB camera, while a FLIR One Pro was used as the thermal camera. The iPhone 15 captured images at a resolution of 1920$\times$1080 at 30 fps, while the FLIR One Pro captured images at a resolution of 160$\times$120 at 8.7 fps. To clearly visualize the temperature distribution, we used the ``grayscale mode" to enhance the visibility of the target objects.

\figA

Using this synchronized dataset, tracking results (bounding boxes, person IDs, confidence scores) were obtained from the teacher model and used as pseudo-labels for thermal images.
\begin{equation}
\bar{y}_{t} = f_{\theta^\text{RGB}}(x_{t}^\text{RGB})
\end{equation}
The student model adapted to thermal images by minimizing a loss function considering both positional and classification errors between its predictions and the teacher model's tracking results.
\begin{equation}
\theta^\text{T} = \arg\min_{\theta} \mathcal{L}^\text{ByteTrack} ( f_{\theta}(x^\text{T}_{1:T}), \bar{y}_{1:T} )
\end{equation}
Furthermore, a fusion model was introduced to complement the characteristics of both the teacher and student models.
\begin{equation}
\hat{y}_{t} = f_{\theta^\text{RGB}\theta^\text{T}}(x_{t}^{\text{T}})
\end{equation}
where $f_{\theta^\text{RGB}\theta^\text{T}}$ represents the fusion function integrating both models. This fusion model integrates the tracking results of the RGB teacher model and the T student model, selecting the result with the higher confidence score in each frame to achieve robust tracking.
For each $i$ $(i \in [1, I^t])$,
$y_t[i] = f^{\text{T}}(x_t^{\text{T}})$,
$\tilde{y}_t[j] = f^{\text{RGB}}(x_t^{\text{T}})$,
and for the corresponding pair $j$ $(j \in [1, I^t])$ satisfying $y_t^\text{ID}[i] = \tilde{y}_t^\text{ID}[j]$, the final fusion result is determined as follows:
\begin{equation}
\hat{y}_t[i] = 
\begin{cases} 
y_t[i] & \text{if } y_t^{\text{SCORE}}[i] > \tilde{y}_t^{\text{SCORE}}[j], \\
\tilde{y}_t[j] & \text{otherwise}.
\end{cases}
\end{equation}

\section{Evaluation Experiment}\label{sec:exp}

The objective of this experiment is to evaluate the performance of RGB-T cooperative vision and identify the optimal cooperation mode.
The acquisition times of input images,
Full=$\{ t \}_{t=1}^T$,
are divided into bright scene times and dark scene times,
which are denoted as
Bright and Dark, respectively
(Bright${\cup}$Dark=Full, Bright${\cap}$Dark=$\emptyset$).
Among all possible combinations of the two scene types (Bright/Dark), two modalities (RGB/T), and three trackers ($f_{\theta^\text{RGB}}$/$f_{\theta^\text{T}}$/$f_{\theta^\text{RGB}\theta^\text{T}}$), meaningless and redundant combinations are excluded based on the following knowledge: (1) No tracker recognized RGB images in dark scenes\footnote{For this reason, all combinations using the RGB modality in dark scenes are excluded.} (2) T images were hardly affected by scene brightness\footnote{For this reason, the performance of \T{T}{Dark} is equivalent to that of \T{T}{Bright}, and the evaluation of \T{T}{Dark} is exempted due to the difficulty in obtaining manual ground-truth annotations.}.
As a result, the following four combinations remain: \RGB{T}{Full}, \T{T}{Full}, \RGB{RGB}{Bright}+\T{T}{Dark}, and {\Fusion}.

This dataset consists of 
a training sequence of 8,970 synchronized RGB thermal image pairs,
and three image sequences obtained from both RGB and thermal cameras in the Bright scene, containing 1,081 frames, 1,049 frames, and 1,010 frames, respectively,
and a single image sequence from  both RGB and thermal cameras in the Dark scene, containing 666 frames.
For performance comparison, the ground truth data is manually annotated with bounding boxes and person IDs.

\subsection{Multi-Person Tracking}

Each MPT tracking method is evaluated using Multiple Object Tracking Accuracy (MOTA) and ID F1 score (IDF1) to assess tracking performance, along with False Positive (FP), True Positive (TP), and False Negative (FN) to evaluate false detections \cite{HOTA}. ID Switches (IDSW) and Higher Order Tracking Accuracy (HOTA) measure trajectory reconstruction, with HOTA providing a global tracking assessment. Unlike MOTA, HOTA balances precision and recall by decomposing detection, assignment, and localization for a more detailed analysis (Table \ref{tab:A}). These metrics gauge how effectively the proposed method tackles key challenges in Multiple Person Tracking (MPT): ID switching, detection failures, occlusion, and handling the appearance/disappearance of individuals. ID switching misassigns identities, detection failures disrupt continuity, and occlusion hides targets. Additionally, newly appearing or disappearing persons may be misidentified, reducing accuracy.

\tabA

\tabB

\RGB{T}{Full} exhibited comparable performance in both Bright and Dark scenes.
However, its performance did not reach that of \RGB{RGB}{Bright} in the Bright scene.

\T{RGB}{Bright}+\T{T}{Dark} achieved comparable performance in both Bright and Dark scenes.

\T{T}{Full} also demonstrated high performance comparable to \T{RGB}{Bright}+\T{T}{Dark} and, surprisingly, matched the performance of \RGB{RGB}{Bright} in the Bright scene (Fig. \ref{fig:A}). 
The concern with this method is that it has the drawback of discarding high-resolution information inherent in RGB images.

{\Fusion}~achieved slightly inferior results compared to \RGB{RGB}{Bright}+\T{T}{Dark}.

The above suggests that the most effective approach among all the combinations considered here is either switching between two trackers using a simple binary brightness classifier that requires RGB input or reusing the RGB tracker as a pseudo-annotation collector.

% Visualization analysis
Figure \ref{fig:DE} visualizes the results of the RGB tracker $f_{\theta^{\text{RGB}}}$ and the T tracker $f_{\theta^{\text{T}}}$ on T camera data.
The RGB tracker $f_{\theta^{\text{RGB}}}$ failed to detect many targets, leading to tracking interruptions and false detections.
In contrast, the T tracker $f_{\theta^{\text{T}}}$, benefiting from knowledge transfer, significantly improved tracking accuracy, reducing missed detections and enhancing ID stability.
This is evident from the visually clearer reconstruction of multiple individuals' boundaries and the effects of occlusion.

\figDE

\subsection{Loop Closure Detection}

In a crowded and dynamic research lab, we conducted a loop closure experiment using a wheeled mobile robot equipped with RGB-T cooperative vision, similar to the setup shown in Figure \ref{fig:B}. The robot utilized HO3 landmarks during the experiment. During this experiment, the robot moved and turned, capturing approximately 500 frames, corresponding to 2-3 minutes. Additionally, multiple humans were present in the scene, with some sitting at desks and working on desktop PCs, while others were seated on a sofa relaxing. In other words, they did not move much. Nevertheless, since the robot was in motion, a challenging scenario arose where multiple humans intersected within the scene.

The research question here is: ``Can HO3 landmarks, useful for loop closing, be extracted from thermal vision?'' We designed a simple image processing algorithm, and despite its simplicity, we found that it is still effective for loop closing.

The loop closure detection method is described below. In the explanation, the x-axis in the image points to the right, and the y-axis points downward. For each image frame:
\begin{itemize}
  \item Detect the human regions in the thermal image.
  \item Detect and track human bounding boxes, assigning person IDs.
  \item Obtain human regions with person IDs by performing pixel-wise AND operations between the bounding box and the human region.
\end{itemize}
For each person ID:
\begin{itemize}
  \item For each x-coordinate within the bounding box, find the pixel belonging to the human region that has the maximum y-coordinate.
\end{itemize}
This set of lower endpoints is referred to as landmarks, which are then recorded in the map along with their frame IDs.
The current frame's landmarks are matched to past landmarks in the map to detect loop closures:
\begin{itemize}
  \item Specifically, given a pair of current and past frames, the matching score from the current frame's landmarks to past landmarks is searched and scored using RANSAC under an affine transformation, then ranked.
\end{itemize}

The performance of loop closure detection was evaluated based on top-5 accuracy. Specifically, in the RANSAC scoring ranking, if the ground-truth match ranked within the top 5, it was considered a success. The frequency of success was then measured and used as the performance metric. The ground-truth frame ID is defined as the frame ID of the viewpoint closest to the robot's trajectory on the viewpoint coordinates.

Performance results are shown in Figure \ref{fig:H}. A meaningful and high loop closure detection performance is achieved. In Figure \ref{fig:F} (b) left, two humans are severely occluded by boxes or partitions. 
Even so, the HO3 landmarks at the boundary between humans and occluders remain highly invariant, contributing to successful loop closure detection. As shown in Figure \ref{fig:F} (b) middle, humans that are not occluded have most of their lower endpoints lying within the bounding box, and in such cases, inappropriate landmarks can be excluded. Figure \ref{fig:F} (b) right shows that the other two HO3 landmarks function as useful landmarks. In this experiment, we demonstrated that HO3 landmarks are effective even compared to the best-known classical change detection and feature extraction methods. Our immediate challenge is to enhance and improve the robustness and accuracy of HO3 landmarks through the introduction of machine learning.

\figH

\figF

\figI

\subsection{Thermal-Depth Cooperative Tracking}\label{sec:td_tracking}

This section evaluates the person tracking performance based on cooperative vision using Thermal (TIR) and Depth images without relying on RGB images.  
The significance of this experiment lies in extending the application scope beyond RGB-T cooperative vision to encompass knowledge transfer among robots equipped with diverse sensor modalities.  
Such efforts represent an important first step toward realizing cross-robot SLAM.  
Comparisons with the RGB-T cooperative vision in the previous section are also conducted to examine the interchangeability of sensor modalities.

The TIR images were acquired by a HIKMICRO Pocket2 thermal camera. Depth images were generated by extracting a narrow field-of-view region corresponding to that of the TIR camera from 3D point clouds captured by a Velodyne VLP-16 LiDAR sensor, and projecting them into image form.

The dataset used in this experiment consists of:

\begin{itemize}
  \item Training set: 6,482 synchronized TIR-Depth image pairs (image resolution: 1440×1080)
  \item Test set: A sequence of 3,797 depth image frames (image resolution: 640×430)
\end{itemize}

All images were manually annotated with person IDs and bounding boxes.

\begin{figure}[t]
\centering
\FIG{8}{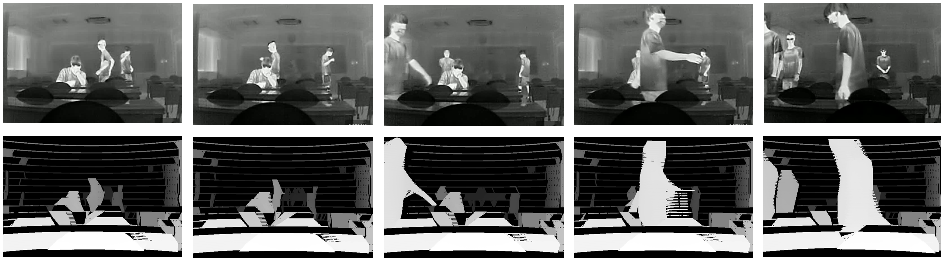}{}\vspace*{0mm}\\
\caption{Cooperative tracking with TIR and depth vision.  
Top row: TIR images,  
Bottom row: Depth images.  
}
\label{fig:K}
\end{figure}

\vspace{1em}
\noindent
Figure~\ref{fig:K} shows paired TIR and depth images. The depth images used here are not derived from high-precision ToF sensors but rather from a very low-resolution subset of a wide-area monitoring 3D LiDAR sensor (Velodyne VLP-16). As a result, fine details such as facial contours are mostly absent, making person identification extremely difficult for human observers.

In fact, many subjects found it easy to distinguish individuals in the TIR images but challenging in the depth images. As illustrated, facial shapes and contours are lost in the depth images.

Nevertheless, our study demonstrates that through knowledge transfer from TIR vision to depth vision, the appearance representation of each person can be effectively complemented, enabling highly accurate person tracking even with low-resolution depth images. This indicates that cooperative vision combined with learning-based knowledge transfer can significantly improve tracking performance in visually challenging environments.

\vspace{1em}
\noindent
Table~\ref{tab:D} presents the performance of the proposed depth-image-based D tracker. In addition to MOTA and IDF1, metrics such as HOTA and ID switch count also show favorable results, confirming the reliable person tracking performance achieved.

\newcommand{\hs}{\hspace*{-1mm}}

\begin{table}[t]
  \centering
  \caption{Person tracking performance with Thermal-Depth cooperative vision (D tracker)}
  \label{tab:D}
  \begin{tabular}{rrrrrrr}
\hline  
MOTA\hs&IDF1\hs&TP\hs&FP\hs&FN\hs&HOTA\hs&IDSW \\ \hline
95.6\hs& 98.0\hs& 10715\hs& 90\hs& 350\hs& 96.1\hs& 48 \\ \hline
  \end{tabular}
\end{table}

\section{Conclusion and Future Works}

In this study, we addressed the cooperative multi-person tracking (MPT) problem using RGB and thermal (T) cameras placed in the same viewpoint --- for the first time.
As expected, the RGB tracker completely failed under the Dark scene's RGB modality and also exhibited degraded performance in thermal imagery.
To address this issue, we confirmed that training the T tracker with pseudo-annotations generated by the RGB tracker is highly effective, achieving state-of-the-art performance in both Bright and Dark scenarios.
We also explored a fusion-based tracker that integrates both RGB and T modalities; surprisingly, however, its performance ranked second.
This suggests that a cooperative strategy --- dynamically switching between RGB and T trackers depending on the situation --- is more effective than a straightforward fusion.
As a demonstration of applicability, we extend our previous work, Human-Object Occlusion Ordering (HO3)-SLAM, to thermal (T) cameras by leveraging the proposed thermal MPT tracker. 
In the experimental section, we present a unique proof-of-concept loop-closure detection experiment in which the human-object-interaction (HOI) between humans and their occluders are treated as pseudo---dynamic landmark feature points (albeit with frequent false negatives).
Whereas conventional SLAM systems typically focus on mapping static landmarks under well-lit conditions, our approach takes a first step toward a new paradigm --- Dynamic-Dark SLAM --- which aims to enable robust mapping across the full spectrum of landmark dynamics (static and dynamic) and lighting conditions (illumination and complete darkness).

\section*{Acknowledgements}

Our work has been supported in part by JSPS KAKENHI Grant-in-Aid for Scientific Research (C) 20K12008 and 23K11270.

\bibliographystyle{unsrt}
\bibliography{reference} 

\end{document}